\documentclass{article}

\usepackage[preprint]{paper}

\usepackage[utf8]{inputenc}
\usepackage[T1]{fontenc}
\usepackage{hyperref}
\usepackage{url}
\usepackage{booktabs}
\usepackage{amsfonts}
\usepackage{nicefrac}
\usepackage{microtype}
\usepackage{xcolor}
\usepackage{adjustbox}
\usepackage{graphicx}
\usepackage{amsmath}
\usepackage{amssymb}
\usepackage{mathtools}
\usepackage{amsthm}
\usepackage{multirow}
\usepackage{colortbl}
\usepackage{bbding}
\usepackage{pifont}
\usepackage{wasysym}
\usepackage{iftex}
\ifPDFTeX
\usepackage{utfsym}
\fi
\usepackage{fontawesome}
\usepackage{natbib}
\setcitestyle{numbers,square}
\hypersetup{
    colorlinks=true,
    linkcolor=blue,
    urlcolor=cyan,
}
\usepackage{arydshln}
\usepackage{array}
\newcolumntype{P}[1]{>{\raggedright\arraybackslash}p{#1}}

\title{An Empirical Study of openPangu Quantization on Ascend NPUs}

\author{\textbf{Tong Shi}$^{1,3}$, \textbf{Jiacheng Wang}$^{1,3}$, \textbf{Hui Xie}$^{1,3}$, \textbf{Ying Li}\textsuperscript{\dag}$^{2}$,\\
\textbf{Aishan Liu}$^{1,2}$, \textbf{Jinyang Guo}$^{1,3}$, \textbf{Xianglong Liu}$^{1,2}$\\
$^{1}$State Key Laboratory of Complex \& Critical Software Environment, Beihang University\\
$^{2}$School of Computer Science and Engineering, Beihang University\\
$^{3}$School of Artificial Intelligence, Beihang University\\
\texttt{\small\{shitong05,jiangchengwang,xiehui,liying,liuaishan,jinyangguo,xlliu\}@buaa.edu.cn}
}

\begin{document}

\maketitle

\begingroup
\renewcommand\thefootnote{\fnsymbol{footnote}}
\footnotetext[2]{Corresponding author.}
\endgroup

\vspace{-0.10in}
\begin{abstract}
openPangu models are attractive targets for private and domestic large-language-model deployment, yet their robustness under aggressive post-training quantization on Ascend NPUs has not been systematically characterized. This paper conducts a controlled empirical study of openPangu 1B and 7B models on Huawei Ascend 910B1 NPUs. We evaluate representative weight-only and weight-activation post-training quantization methods, including RTN, GPTQ, AWQ, SmoothQuant, GPTAQ, BiLLM, and SliM-LLM, under a unified calibration and evaluation protocol.
Across 18 evaluation tasks, we find that 8-bit weight-only quantization is effectively lossless for both models, while 4-bit quantization remains practical for the 7B model but is visibly more harmful for the 1B model on reasoning, math, and code tasks. Ultra-low precision remains challenging: most 2-bit and binary settings collapse to near-random behavior, and W4A4 SmoothQuant produces non-finite perplexity in our evaluation. These results provide an NPU-oriented accuracy map for selecting openPangu quantization settings and highlight the persistent difficulty of extreme low-bit compression.
\vspace{-0.10in}
\end{abstract}

\section{Introduction}

Large language models based on the Transformer architecture~\cite{vaswani2017attention} are increasingly deployed in private, domain-specific, and hardware-constrained environments where inference cost is as important as model quality. openPangu is a representative domestic autoregressive LLM family for such scenarios. Compared with CUDA-centered deployment studies, openPangu deployment on Ascend NPUs introduces a different systems context, including NPU-oriented operator support, device memory constraints, and framework compatibility. These factors make it necessary to re-examine whether commonly used quantization recipes remain reliable under the target deployment stack.

Post-training quantization (PTQ) is a direct way to reduce memory footprint and enable larger models or larger batches under fixed HBM budgets~\cite{gong2024survey}. Mature LLM PTQ methods such as GPTQ~\cite{frantar2022gptq}, AWQ~\cite{lin2023awq}, SmoothQuant~\cite{xiao2023smoothquant}, and BiLLM~\cite{huang2024billm} have been widely studied, but their accuracy behavior is strongly model- and implementation-dependent. For openPangu on Ascend NPUs, the central question is not whether a method performs well on a public CUDA benchmark, but which bit-widths remain credible under a consistent NPU-oriented calibration and evaluation protocol.

In this empirical study, we systematically evaluate openPangu 1B and 7B under a broad set of PTQ settings. We test RTN, GPTQ, AWQ, SmoothQuant, GPTAQ, BiLLM, and SliM-LLM, spanning bit-widths from 1 to 8 bits. Our evaluation covers language modeling perplexity on WikiText2~\cite{merity2016pointer} and C4~\cite{raffel2020exploring}, English commonsense reasoning tasks including PIQA~\cite{bisk2020piqa}, ARC~\cite{clark2018think}, HellaSwag~\cite{zellers2019hellaswag}, Winogrande~\cite{sakaguchi2021winogrande}, and BoolQ~\cite{boolq}, as well as MMLU~\cite{hendrycks2020measuring}, Chinese knowledge benchmarks, mathematical reasoning, reading comprehension, code generation, and instruction following. This study aims to: (1) benchmark the quantization-induced accuracy trade-offs of openPangu on Ascend NPUs, (2) identify reliable bit-width and method choices for practical deployment, and (3) highlight unresolved challenges in ultra-low-bit quantization.

\section{Empirical Study}

\subsection{Experiment Settings}

We evaluate two openPangu autoregressive language models, openPangu-1B and openPangu-7B. Both models use the \texttt{PanguEmbedded} architecture and are evaluated in HuggingFace format. Table~\ref{tab:model-config} summarizes their main architectural configurations.

\begin{table}[t]
\centering
\small
\caption{openPangu model configurations used in this study.}
\begin{tabular}{lcccccc}
\hline
\textbf{Model} & \textbf{Layers} & \textbf{Hidden} & \textbf{FFN} & \textbf{Q heads} & \textbf{KV heads} & \textbf{Context} \\
\hline
openPangu-1B & 26 & 1536 & 6144 & 12 & 6 & 32768 \\
openPangu-7B & 34 & 4096 & 12800 & 32 & 8 & 32768 \\
\hline
\end{tabular}
\label{tab:model-config}
\end{table}

Experiments are conducted on Huawei Ascend 910B1 NPUs with 64GB HBM per device. Each quantization or evaluation run uses one NPU to ensure consistent device conditions. The software stack is based on PyTorch 2.5.1, torch-npu 2.5.1, Transformers 4.53.2, and lm-evaluation-harness. Quantization is implemented with LightCompress/LLMC-compatible, BiLLM-compatible, and SliM-LLM-compatible flows.

\textbf{Quantization methods.}
To comprehensively assess openPangu's quantization robustness, we select representative PTQ methods covering different technical families: Round-To-Nearest (RTN), GPTQ~\cite{frantar2022gptq}, AWQ~\cite{lin2023awq}, SmoothQuant~\cite{xiao2023smoothquant}, GPTAQ~\cite{li2025gptaq}, BiLLM~\cite{huang2024billm}, and SliM-LLM~\cite{huang2025slimllm}. RTN, GPTQ, AWQ, and GPTAQ are evaluated as weight-only quantization methods. SmoothQuant is evaluated as a weight-activation quantization method. BiLLM and SliM-LLM represent binary and 2-bit ultra-low-precision settings.

\textbf{Quantization protocol.}
To ensure fair comparison across methods, we keep the calibration and grouping protocol fixed whenever calibration is required. Calibrated experiments use 128 Alpaca-style instruction samples with sequence length 512. Weight-only methods use symmetric per-group quantization with group size 128 and activation bit-width fixed to 16. GPTQ uses act-order, 0.01 damping, block size 128, and true sequential quantization. AWQ uses transformation and weight clipping. GPTAQ follows GPTQ-style settings with an activation-aware scaling coefficient. SmoothQuant evaluates W8A8 and W4A4 settings with per-channel weight quantization, per-token activation quantization, and smoothing coefficient 0.8. RTN does not use calibration data.

\textbf{Evaluation protocol.}
For a comprehensive PTQ evaluation, we measure perplexity (PPL) on WikiText2~\cite{merity2016pointer} and C4~\cite{raffel2020exploring}. We evaluate English commonsense reasoning on PIQA~\cite{bisk2020piqa}, Winogrande~\cite{sakaguchi2021winogrande}, ARC-Easy, ARC-Challenge~\cite{clark2018think}, HellaSwag~\cite{zellers2019hellaswag}, and BoolQ~\cite{boolq}. We further evaluate broad knowledge and reasoning with MMLU~\cite{hendrycks2020measuring}, CMMLU, C-Eval, CLUEWSC, GSM8K, MATH-500, DROP, MBPP, HumanEval, and IF-Eval. Multiple-choice tasks use accuracy or normalized accuracy following their lm-evaluation-harness defaults. GSM8K uses exact match, MATH-500 uses math verification, DROP uses F1, and code-generation tasks use pass@1. Non-finite perplexity values are reported directly, and failed runs are marked as ``Failed''.

\subsection{PTQ Results}

We present detailed experimental results in Table~\ref{tab:openpangu-1b-main}, Table~\ref{tab:openpangu-1b-extended}, Table~\ref{tab:openpangu-7b-main}, and Table~\ref{tab:openpangu-7b-extended}. Following common PTQ reporting practice, \#W denotes the weight quantization bit-width, \#A denotes the activation quantization bit-width, and \#G denotes the group size. PPL metrics are lower-is-better, while all accuracy, F1, pass@1, and instruction-following metrics are higher-is-better.

\begin{table*}[p]
\setlength{\tabcolsep}{1.2pt}
\renewcommand{\arraystretch}{1.30}
\footnotesize
\centering
\caption{PTQ results of openPangu-1B on language modeling, English commonsense reasoning, and MMLU. We report PPL on WikiText2 and C4, six zero-shot commonsense tasks, their average accuracy, and MMLU. \#W denotes weight bit-width, \#A denotes activation bit-width, and \#G denotes group size.}
\begin{adjustbox}{max width=\linewidth}
\begin{tabular}{ll||ccc|cc|ccccccc|c}
\hline
\textbf{Model} & \textbf{Method} & \textbf{\#W} & \textbf{\#A} & \textbf{\#G} & \textbf{Wiki2$(\downarrow)$} & \textbf{C4$(\downarrow)$} & \textbf{PIQA} & \textbf{ARC-E} & \textbf{ARC-C} & \textbf{HellaS} & \textbf{WinoG} & \textbf{BoolQ} & \textbf{Avg$(\uparrow)$} & \textbf{MMLU} \\
\hline
\multirow{37}{*}{\centering openPangu-1B} & FP16 & 16 & 16 & / & 25.72 & 75.46 & 69.97 & 57.87 & 33.62 & 59.06 & 56.51 & 62.54 & 56.59 & 48.88\\
\cdashline{2-15}
& RTN & 8 & 16 & 128 & 25.81 & 75.82 & 69.70 & 58.00 & 33.96 & 59.27 & 56.99 & 60.95 & 56.48 & 48.92\\
& AWQ & 8 & 16 & 128 & 25.63 & 75.44 & 69.86 & 58.00 & 34.22 & 59.07 & 57.70 & 63.00 & 56.98 & 49.14\\
& GPTQ & 8 & 16 & 128 & 25.73 & 75.51 & 69.80 & 58.46 & 33.62 & 59.21 & 56.75 & 63.00 & 56.81 & 48.75\\
& GPTAQ & 8 & 16 & 128 & 27.08 & 75.79 & 69.70 & 58.25 & 33.70 & 59.35 & 56.75 & 62.60 & 56.73 & 48.24\\
& SmoothQuant & 8 & 8 & / & 26.17 & inf & 70.35 & 57.91 & 34.47 & 58.72 & 58.96 & 64.65 & 57.51 & 46.18\\
\cdashline{2-15}
& RTN & 7 & 16 & 128 & 26.00 & 75.54 & 70.08 & 58.38 & 34.56 & 59.27 & 57.38 & 63.88 & 57.26 & 48.89\\
& AWQ & 7 & 16 & 128 & 26.17 & 75.97 & 70.08 & 58.33 & 34.22 & 59.04 & 56.99 & 63.76 & 57.07 & 49.00\\
& GPTQ & 7 & 16 & 128 & 25.77 & 76.27 & 70.02 & 58.80 & 33.45 & 59.10 & 56.67 & 61.96 & 56.67 & 48.56\\
& GPTAQ & 7 & 16 & 128 & 27.21 & 76.40 & 70.51 & 58.04 & 34.30 & 59.19 & 56.51 & 63.12 & 56.95 & 48.75\\
\cdashline{2-15}
& RTN & 6 & 16 & 128 & 27.03 & 79.42 & 69.80 & 57.45 & 34.30 & 58.65 & 57.38 & 64.43 & 57.00 & 47.90\\
& AWQ & 6 & 16 & 128 & 25.97 & 76.03 & 70.18 & 58.12 & 33.62 & 59.19 & 57.30 & 65.20 & 57.27 & 47.93\\
& GPTQ & 6 & 16 & 128 & 25.74 & 76.77 & 70.84 & 58.59 & 34.39 & 58.84 & 55.96 & 64.74 & 57.23 & 47.68\\
& GPTAQ & 6 & 16 & 128 & 27.34 & 77.22 & 70.13 & 54.67 & 33.19 & 58.53 & 56.59 & 60.49 & 55.60 & 47.31\\
\cdashline{2-15}
& RTN & 5 & 16 & 128 & 28.26 & 84.40 & 68.23 & 57.95 & 32.85 & 57.26 & 56.20 & 55.90 & 54.73 & 46.45\\
& AWQ & 5 & 16 & 128 & 27.19 & 79.77 & 70.51 & 55.77 & 33.79 & 58.35 & 57.14 & 56.91 & 55.41 & 48.23\\
& GPTQ & 5 & 16 & 128 & 29.29 & 82.37 & 69.59 & 56.65 & 34.04 & 57.85 & 57.06 & 66.12 & 56.88 & 46.96\\
& GPTAQ & 5 & 16 & 128 & 29.25 & 81.87 & 69.80 & 57.62 & 34.81 & 58.36 & 57.85 & 64.28 & 57.12 & 45.09\\
\cdashline{2-15}
& RTN & 4 & 16 & 128 & 43.30 & 124.86 & 64.58 & 47.52 & 33.62 & 52.57 & 53.28 & 68.07 & 53.27 & 31.37\\
& AWQ & 4 & 16 & 128 & 38.58 & 98.98 & 67.79 & 54.67 & 33.62 & 55.34 & 57.93 & 63.09 & 55.41 & 41.08\\
& GPTQ & 4 & 16 & 128 & 48.82 & 119.26 & 65.83 & 53.41 & 33.19 & 54.51 & 55.09 & 60.80 & 53.80 & 40.02\\
& GPTAQ & 4 & 16 & 128 & 45.23 & 110.72 & 67.25 & 52.74 & 31.48 & 53.80 & 56.04 & 57.89 & 53.20 & 38.69\\
& SmoothQuant & 4 & 4 & / & inf & inf & 50.76 & 26.43 & 25.94 & 25.43 & 47.28 & 42.29 & 36.35 & 23.94\\
\cdashline{2-15}
& RTN & 3 & 16 & 128 & 4301.15 & 6128.22 & 55.17 & 33.50 & 24.40 & 30.51 & 49.17 & 46.54 & 39.88 & 23.23\\
& AWQ & 3 & 16 & 128 & 171.92 & 395.02 & 61.04 & 44.44 & 29.27 & 41.22 & 52.57 & 64.50 & 48.84 & 30.03\\
& GPTQ & 3 & 16 & 128 & 7338.89 & 5959.38 & 54.35 & 32.49 & 26.88 & 32.12 & 52.25 & 50.64 & 41.45 & 24.55\\
& GPTAQ & 3 & 16 & 128 & 8667.14 & 4110.11 & 55.71 & 32.87 & 23.21 & 32.71 & 49.88 & 53.73 & 41.35 & 23.84\\
\cdashline{2-15}
& RTN & 2 & 16 & 128 & 2.42e+06 & 1.92e+07 & 49.46 & 27.02 & 26.19 & 25.87 & 50.28 & 37.83 & 36.11 & 25.35\\
& AWQ & 2 & 16 & 128 & 199046.21 & 337551.72 & 49.62 & 26.56 & 26.45 & 26.18 & 51.30 & 40.86 & 36.83 & 24.85\\
& GPTQ & 2 & 16 & 128 & 425987.81 & 3.19e+06 & 50.92 & 25.04 & 27.65 & 25.82 & 51.54 & 45.87 & 37.81 & 24.15\\
& GPTAQ & 2 & 16 & 128 & 555055.32 & 2.40e+06 & 51.36 & 25.80 & 27.99 & 25.97 & 49.41 & 43.00 & 37.26 & 25.51\\
& SliM-LLM & 2 & 16 & 128 & 2.26e+06 & 4.39e+07 & 52.99 & 25.13 & 26.54 & 26.20 & 47.59 & 44.07 & 37.09 & 24.70\\
\cdashline{2-15}
& RTN & 1 & 16 & 128 & NaN & NaN & 49.51 & 25.08 & 22.70 & 25.04 & 49.57 & 37.83 & 34.95 & 22.95\\
& AWQ & 1 & 16 & 128 & Failed & Failed & Failed & Failed & Failed & Failed & Failed & Failed & Failed & Failed\\
& GPTQ & 1 & 16 & 128 & NaN & NaN & 49.51 & 25.08 & 22.70 & 25.04 & 49.57 & 37.83 & 34.95 & 22.95\\
& GPTAQ & 1 & 16 & 128 & NaN & NaN & 49.51 & 25.08 & 22.70 & 25.04 & 49.57 & 37.83 & 34.95 & 22.95\\
& BiLLM & 1 & 16 & 128 & 1.22e+06 & 6.93e+06 & 49.89 & 25.67 & 27.47 & 26.80 & 51.14 & 37.95 & 36.49 & 25.79\\
\hline
\end{tabular}
\end{adjustbox}
\label{tab:openpangu-1b-main}
\end{table*}

\begin{table*}[p]
\setlength{\tabcolsep}{1.2pt}
\renewcommand{\arraystretch}{1.30}
\footnotesize
\centering
\caption{Extended PTQ results of openPangu-1B on Chinese knowledge, math, reading, code, and instruction-following tasks. Values are percentages except DROP F1.}
\begin{adjustbox}{max width=\linewidth}
\begin{tabular}{ll||ccc|ccccccccc}
\hline
\textbf{Model} & \textbf{Method} & \textbf{\#W} & \textbf{\#A} & \textbf{\#G} & \textbf{CMMLU} & \textbf{C-Eval} & \textbf{CLUEWSC} & \textbf{GSM8K} & \textbf{MATH} & \textbf{DROP} & \textbf{MBPP} & \textbf{HumanEval} & \textbf{IF-Eval} \\
\hline
\multirow{37}{*}{\centering openPangu-1B} & FP16 & 16 & 16 & / & 44.31 & 43.98 & 62.83 & 30.10 & 23.20 & 8.17 & 21.60 & 20.12 & 28.47\\
\cdashline{2-14}
& RTN & 8 & 16 & 128 & 43.68 & 42.64 & 65.13 & 30.93 & 24.80 & 8.10 & 19.00 & 21.95 & 29.39\\
& AWQ & 8 & 16 & 128 & 43.84 & 42.94 & 63.49 & 29.19 & 23.40 & 8.06 & 20.40 & 21.34 & 27.54\\
& GPTQ & 8 & 16 & 128 & 44.80 & 45.17 & 66.78 & 29.87 & 24.60 & 8.23 & 19.60 & 19.51 & 29.76\\
& GPTAQ & 8 & 16 & 128 & 44.78 & 43.31 & 65.46 & 30.17 & 24.20 & 8.14 & 18.60 & 19.51 & 27.54\\
& SmoothQuant & 8 & 8 & / & 42.48 & 37.07 & 57.89 & 28.66 & 24.80 & 7.85 & 20.60 & 18.29 & 28.84\\
\cdashline{2-14}
& RTN & 7 & 16 & 128 & 44.07 & 43.16 & 60.86 & 31.54 & 29.60 & 8.23 & 21.60 & 21.34 & 29.57\\
& AWQ & 7 & 16 & 128 & 44.22 & 43.24 & 63.82 & 29.80 & 24.00 & 8.33 & 18.20 & 20.73 & 27.91\\
& GPTQ & 7 & 16 & 128 & 43.82 & 43.46 & 50.00 & 32.37 & 23.40 & 8.03 & 21.40 & 19.51 & 29.39\\
& GPTAQ & 7 & 16 & 128 & 43.47 & 44.58 & 53.62 & 29.26 & 27.00 & 7.91 & 18.60 & 19.51 & 28.28\\
\cdashline{2-14}
& RTN & 6 & 16 & 128 & 44.51 & 48.89 & 63.82 & 28.81 & 25.40 & 7.11 & 10.00 & 18.29 & 29.39\\
& AWQ & 6 & 16 & 128 & 43.62 & 44.21 & 63.16 & 29.80 & 23.80 & 8.51 & 20.00 & 18.90 & 26.25\\
& GPTQ & 6 & 16 & 128 & 41.88 & 43.31 & 66.12 & 29.11 & 26.80 & 7.83 & 22.20 & 20.73 & 28.84\\
& GPTAQ & 6 & 16 & 128 & 42.24 & 42.64 & 64.14 & 29.87 & 24.20 & 8.26 & 21.00 & 18.29 & 29.39\\
\cdashline{2-14}
& RTN & 5 & 16 & 128 & 42.84 & 41.23 & 69.41 & 26.61 & 18.00 & 6.87 & 19.20 & 15.85 & 27.36\\
& AWQ & 5 & 16 & 128 & 43.05 & 41.23 & 69.08 & 27.22 & 19.80 & 7.36 & 14.40 & 18.90 & 26.80\\
& GPTQ & 5 & 16 & 128 & 43.14 & 46.29 & 59.54 & 27.98 & 22.60 & 8.32 & 20.60 & 18.29 & 27.73\\
& GPTAQ & 5 & 16 & 128 & 40.66 & 44.87 & 67.76 & 24.26 & 20.60 & 7.86 & 20.40 & 18.29 & 25.88\\
\cdashline{2-14}
& RTN & 4 & 16 & 128 & 32.80 & 34.62 & 46.71 & 9.55 & 9.40 & 4.09 & 0.20 & 6.71 & 17.74\\
& AWQ & 4 & 16 & 128 & 31.39 & 29.12 & 46.05 & 10.01 & 11.20 & 4.72 & 6.20 & 9.15 & 22.92\\
& GPTQ & 4 & 16 & 128 & 36.43 & 40.04 & 49.67 & 16.53 & 9.20 & 5.82 & 7.80 & 4.27 & 24.77\\
& GPTAQ & 4 & 16 & 128 & 37.24 & 38.41 & 41.78 & 13.72 & 9.40 & 5.33 & 3.40 & 4.88 & 20.70\\
& SmoothQuant & 4 & 4 & / & 25.19 & 24.22 & 36.84 & 0.30 & 0.00 & 0.09 & 0.00 & 0.00 & 6.47\\
\cdashline{2-14}
& RTN & 3 & 16 & 128 & 24.71 & 26.30 & 36.51 & 1.29 & 0.20 & 0.53 & 0.00 & 0.00 & 11.09\\
& AWQ & 3 & 16 & 128 & 28.23 & 26.45 & 36.51 & 2.20 & 2.80 & 2.62 & 0.20 & 0.00 & 15.53\\
& GPTQ & 3 & 16 & 128 & 24.52 & 24.37 & 37.17 & 0.99 & 2.00 & 0.62 & 0.00 & 0.00 & 9.24\\
& GPTAQ & 3 & 16 & 128 & 25.21 & 24.15 & 37.17 & 1.14 & 0.60 & 0.57 & 0.00 & 0.00 & 10.35\\
\cdashline{2-14}
& RTN & 2 & 16 & 128 & 25.59 & 24.22 & 36.51 & 0.08 & 0.00 & 0.07 & 0.00 & 0.00 & 8.32\\
& AWQ & 2 & 16 & 128 & 24.49 & 24.37 & 36.51 & 0.00 & 0.00 & 0.13 & 0.00 & 0.00 & 8.69\\
& GPTQ & 2 & 16 & 128 & 24.73 & 26.00 & 36.51 & 0.91 & 2.00 & 0.21 & 0.00 & 0.00 & 7.76\\
& GPTAQ & 2 & 16 & 128 & 24.87 & 24.96 & 36.84 & 0.76 & 3.00 & 0.23 & 0.00 & 0.00 & 9.98\\
& SliM-LLM & 2 & 16 & 128 & 25.34 & 25.71 & 36.18 & 0.08 & 0.60 & 0.12 & 0.00 & 0.00 & 7.39\\
\cdashline{2-14}
& RTN & 1 & 16 & 128 & 25.26 & 23.03 & 36.51 & 0.00 & 0.00 & 0.00 & 0.00 & 0.00 & 0.00\\
& AWQ & 1 & 16 & 128 & Failed & Failed & Failed & Failed & Failed & Failed & Failed & Failed & Failed\\
& GPTQ & 1 & 16 & 128 & 25.26 & 23.03 & 36.51 & 0.00 & 0.00 & 0.00 & 0.00 & 0.00 & 0.00\\
& GPTAQ & 1 & 16 & 128 & 25.26 & 23.03 & 36.51 & 0.00 & 0.00 & 0.00 & 0.00 & 0.00 & 0.00\\
& BiLLM & 1 & 16 & 128 & 25.00 & 23.63 & 36.51 & 0.00 & 0.00 & 0.16 & 0.00 & 0.00 & 10.91\\
\hline
\end{tabular}
\end{adjustbox}
\label{tab:openpangu-1b-extended}
\end{table*}

\begin{table*}[p]
\setlength{\tabcolsep}{1.2pt}
\renewcommand{\arraystretch}{1.30}
\footnotesize
\centering
\caption{PTQ results of openPangu-7B on language modeling, English commonsense reasoning, and MMLU. We report PPL on WikiText2 and C4, six zero-shot commonsense tasks, their average accuracy, and MMLU.}
\begin{adjustbox}{max width=\linewidth}
\begin{tabular}{ll||ccc|cc|ccccccc|c}
\hline
\textbf{Model} & \textbf{Method} & \textbf{\#W} & \textbf{\#A} & \textbf{\#G} & \textbf{Wiki2$(\downarrow)$} & \textbf{C4$(\downarrow)$} & \textbf{PIQA} & \textbf{ARC-E} & \textbf{ARC-C} & \textbf{HellaS} & \textbf{WinoG} & \textbf{BoolQ} & \textbf{Avg$(\uparrow)$} & \textbf{MMLU} \\
\hline
\multirow{37}{*}{\centering openPangu-7B} & FP16 & 16 & 16 & / & 15.79 & 80.43 & 71.44 & 62.12 & 45.82 & 73.50 & 64.88 & 80.83 & 66.43 & 68.22\\
\cdashline{2-15}
& RTN & 8 & 16 & 128 & 15.79 & 80.48 & 70.84 & 61.95 & 45.05 & 73.36 & 64.25 & 80.89 & 66.06 & 68.20\\
& AWQ & 8 & 16 & 128 & 15.74 & 80.27 & 71.65 & 61.91 & 45.56 & 73.46 & 64.96 & 81.07 & 66.44 & 68.20\\
& GPTQ & 8 & 16 & 128 & 15.84 & 80.64 & 71.11 & 61.70 & 45.90 & 73.40 & 65.51 & 80.83 & 66.41 & 67.88\\
& GPTAQ & 8 & 16 & 128 & 16.42 & 80.73 & 71.11 & 61.62 & 45.31 & 73.41 & 65.04 & 81.10 & 66.27 & 68.17\\
& SmoothQuant & 8 & 8 & / & 15.98 & 81.30 & 69.75 & 62.16 & 43.34 & 73.50 & 64.72 & 81.80 & 65.88 & 68.68\\
\cdashline{2-15}
& RTN & 7 & 16 & 128 & 15.82 & 79.38 & 71.22 & 61.03 & 45.14 & 73.29 & 65.75 & 80.34 & 66.13 & 67.89\\
& AWQ & 7 & 16 & 128 & 15.84 & 80.60 & 71.49 & 62.29 & 45.31 & 73.29 & 65.19 & 80.89 & 66.41 & 68.12\\
& GPTQ & 7 & 16 & 128 & 15.66 & 80.67 & 71.22 & 61.87 & 44.88 & 73.21 & 64.56 & 80.55 & 66.05 & 68.47\\
& GPTAQ & 7 & 16 & 128 & 16.20 & 80.54 & 71.87 & 62.75 & 46.33 & 73.20 & 64.56 & 79.85 & 66.43 & 68.02\\
\cdashline{2-15}
& RTN & 6 & 16 & 128 & 15.67 & 75.75 & 72.03 & 62.16 & 44.20 & 73.54 & 64.88 & 81.31 & 66.35 & 69.50\\
& AWQ & 6 & 16 & 128 & 15.88 & 80.58 & 72.25 & 62.42 & 43.60 & 73.28 & 65.04 & 81.59 & 66.36 & 68.52\\
& GPTQ & 6 & 16 & 128 & 16.45 & 83.09 & 71.65 & 62.96 & 43.52 & 72.79 & 66.14 & 77.68 & 65.79 & 69.48\\
& GPTAQ & 6 & 16 & 128 & 16.66 & 82.94 & 70.95 & 62.50 & 45.05 & 73.53 & 64.40 & 79.02 & 65.91 & 68.57\\
\cdashline{2-15}
& RTN & 5 & 16 & 128 & 15.67 & 77.49 & 70.67 & 63.13 & 44.88 & 73.48 & 64.33 & 80.80 & 66.22 & 70.15\\
& AWQ & 5 & 16 & 128 & 15.90 & 79.92 & 72.36 & 62.63 & 44.11 & 73.05 & 65.27 & 78.81 & 66.04 & 67.81\\
& GPTQ & 5 & 16 & 128 & 16.75 & 85.05 & 72.36 & 64.69 & 43.00 & 72.35 & 62.12 & 79.79 & 65.72 & 69.01\\
& GPTAQ & 5 & 16 & 128 & 17.05 & 82.99 & 69.64 & 62.04 & 41.72 & 72.13 & 61.88 & 78.44 & 64.31 & 68.96\\
\cdashline{2-15}
& RTN & 4 & 16 & 128 & 18.19 & 86.42 & 70.95 & 60.90 & 41.04 & 69.71 & 64.17 & 82.45 & 64.87 & 65.18\\
& AWQ & 4 & 16 & 128 & 16.91 & 80.60 & 70.18 & 60.40 & 41.98 & 71.15 & 65.19 & 81.99 & 65.15 & 63.45\\
& GPTQ & 4 & 16 & 128 & 20.99 & 97.02 & 69.80 & 60.19 & 40.27 & 68.96 & 62.27 & 79.88 & 63.56 & 63.91\\
& GPTAQ & 4 & 16 & 128 & 21.16 & 94.96 & 67.19 & 56.65 & 39.59 & 68.20 & 59.98 & 79.08 & 61.78 & 62.01\\
& SmoothQuant & 4 & 4 & / & inf & inf & 49.84 & 26.52 & 26.02 & 25.93 & 49.41 & 39.14 & 36.14 & 25.20\\
\cdashline{2-15}
& RTN & 3 & 16 & 128 & 41.00 & 220.73 & 61.10 & 43.94 & 32.08 & 52.14 & 54.62 & 66.54 & 51.74 & 36.60\\
& AWQ & 3 & 16 & 128 & 26.42 & 105.99 & 68.72 & 63.05 & 39.85 & 64.30 & 61.96 & 76.91 & 62.46 & 51.90\\
& GPTQ & 3 & 16 & 128 & 133.43 & 1098.20 & 54.08 & 31.10 & 26.28 & 37.03 & 53.12 & 54.40 & 42.67 & 25.61\\
& GPTAQ & 3 & 16 & 128 & 148.24 & 1111.15 & 54.24 & 31.65 & 27.47 & 35.30 & 51.46 & 57.37 & 42.91 & 24.48\\
\cdashline{2-15}
& RTN & 2 & 16 & 128 & 8.02e+06 & 8.35e+07 & 50.33 & 25.76 & 26.79 & 26.49 & 48.86 & 48.47 & 37.78 & 23.82\\
& AWQ & 2 & 16 & 128 & 616119.18 & 1.14e+06 & 51.14 & 27.48 & 25.17 & 26.06 & 49.57 & 44.16 & 37.26 & 23.30\\
& GPTQ & 2 & 16 & 128 & 2.13e+07 & 2.18e+08 & 51.85 & 24.96 & 29.18 & 25.58 & 51.22 & 49.63 & 38.74 & 24.65\\
& GPTAQ & 2 & 16 & 128 & 3.01e+07 & 2.49e+08 & 50.05 & 25.63 & 27.56 & 26.63 & 51.46 & 49.60 & 38.49 & 24.55\\
& SliM-LLM & 2 & 16 & 128 & 67152.91 & 1.83e+06 & 50.33 & 25.13 & 27.05 & 25.99 & 49.33 & 41.01 & 36.47 & 24.49\\
\cdashline{2-15}
& RTN & 1 & 16 & 128 & NaN & NaN & 49.51 & 25.08 & 22.70 & 25.04 & 49.57 & 37.83 & 34.95 & 22.95\\
& AWQ & 1 & 16 & 128 & Failed & Failed & Failed & Failed & Failed & Failed & Failed & Failed & Failed & Failed\\
& GPTQ & 1 & 16 & 128 & NaN & NaN & 49.51 & 25.08 & 22.70 & 25.04 & 49.57 & 37.83 & 34.95 & 22.95\\
& GPTAQ & 1 & 16 & 128 & NaN & NaN & 49.51 & 25.08 & 22.70 & 25.04 & 49.57 & 37.83 & 34.95 & 22.95\\
& BiLLM & 1 & 16 & 128 & 1625.24 & 12688.57 & 52.34 & 29.04 & 21.84 & 27.19 & 51.62 & 59.02 & 40.18 & 23.89\\
\hline
\end{tabular}
\end{adjustbox}
\label{tab:openpangu-7b-main}
\end{table*}

\begin{table*}[p]
\setlength{\tabcolsep}{1.2pt}
\renewcommand{\arraystretch}{1.30}
\footnotesize
\centering
\caption{Extended PTQ results of openPangu-7B on Chinese knowledge, math, reading, code, and instruction-following tasks. Values are percentages except DROP F1.}
\begin{adjustbox}{max width=\linewidth}
\begin{tabular}{ll||ccc|ccccccccc}
\hline
\textbf{Model} & \textbf{Method} & \textbf{\#W} & \textbf{\#A} & \textbf{\#G} & \textbf{CMMLU} & \textbf{C-Eval} & \textbf{CLUEWSC} & \textbf{GSM8K} & \textbf{MATH} & \textbf{DROP} & \textbf{MBPP} & \textbf{HumanEval} & \textbf{IF-Eval} \\
\hline
\multirow{37}{*}{\centering openPangu-7B} & FP16 & 16 & 16 & / & 63.47 & 66.57 & 36.51 & 64.14 & 44.80 & 5.59 & 37.80 & 40.85 & 33.27\\
\cdashline{2-14}
& RTN & 8 & 16 & 128 & 63.43 & 66.57 & 36.51 & 66.03 & 46.20 & 5.59 & 37.80 & 40.24 & 34.38\\
& AWQ & 8 & 16 & 128 & 63.38 & 65.75 & 36.51 & 65.28 & 44.80 & 5.53 & 37.40 & 41.46 & 34.38\\
& GPTQ & 8 & 16 & 128 & 63.44 & 65.82 & 36.51 & 62.85 & 45.00 & 5.47 & 37.40 & 39.63 & 34.57\\
& GPTAQ & 8 & 16 & 128 & 63.09 & 65.90 & 36.51 & 63.31 & 42.20 & 5.57 & 37.80 & 42.07 & 32.72\\
& SmoothQuant & 8 & 8 & / & 62.97 & 66.86 & 37.83 & 67.78 & 44.20 & 5.85 & 32.20 & 31.71 & 32.72\\
\cdashline{2-14}
& RTN & 7 & 16 & 128 & 63.05 & 65.90 & 36.84 & 65.81 & 48.40 & 5.56 & 38.20 & 36.59 & 32.90\\
& AWQ & 7 & 16 & 128 & 63.71 & 67.09 & 36.84 & 62.77 & 43.60 & 5.49 & 37.40 & 42.68 & 33.64\\
& GPTQ & 7 & 16 & 128 & 62.80 & 65.30 & 37.17 & 62.40 & 40.00 & 5.52 & 33.80 & 35.98 & 34.01\\
& GPTAQ & 7 & 16 & 128 & 63.79 & 66.64 & 36.84 & 66.64 & 43.40 & 5.71 & 38.60 & 39.02 & 34.75\\
\cdashline{2-14}
& RTN & 6 & 16 & 128 & 62.93 & 65.97 & 36.51 & 74.68 & 46.40 & 5.14 & 39.40 & 33.54 & 33.64\\
& AWQ & 6 & 16 & 128 & 64.08 & 67.01 & 36.84 & 60.80 & 44.20 & 5.48 & 37.40 & 39.63 & 32.35\\
& GPTQ & 6 & 16 & 128 & 62.69 & 65.60 & 36.84 & 65.13 & 43.20 & 5.38 & 28.40 & 29.27 & 34.57\\
& GPTAQ & 6 & 16 & 128 & 64.43 & 65.30 & 36.84 & 67.10 & 44.80 & 5.54 & 32.20 & 35.37 & 32.53\\
\cdashline{2-14}
& RTN & 5 & 16 & 128 & 63.37 & 68.80 & 36.51 & 70.20 & 41.40 & 5.39 & 34.80 & 31.71 & 31.24\\
& AWQ & 5 & 16 & 128 & 60.15 & 63.00 & 36.51 & 69.37 & 44.80 & 5.62 & 39.20 & 44.51 & 30.50\\
& GPTQ & 5 & 16 & 128 & 62.93 & 67.46 & 45.07 & 68.76 & 42.20 & 5.27 & 38.20 & 20.73 & 34.01\\
& GPTAQ & 5 & 16 & 128 & 63.69 & 67.90 & 45.07 & 69.37 & 41.40 & 5.57 & 36.20 & 23.78 & 33.09\\
\cdashline{2-14}
& RTN & 4 & 16 & 128 & 52.32 & 58.25 & 41.78 & 68.69 & 41.60 & 6.37 & 28.60 & 33.54 & 33.64\\
& AWQ & 4 & 16 & 128 & 61.03 & 62.93 & 42.11 & 69.83 & 43.40 & 5.63 & 39.60 & 33.54 & 34.01\\
& GPTQ & 4 & 16 & 128 & 53.57 & 55.57 & 48.36 & 58.98 & 32.40 & 4.48 & 28.60 & 9.76 & 32.53\\
& GPTAQ & 4 & 16 & 128 & 52.24 & 54.90 & 38.49 & 57.54 & 31.60 & 5.03 & 26.00 & 13.41 & 30.68\\
& SmoothQuant & 4 & 4 & / & 24.49 & 21.47 & 36.51 & 0.08 & 0.00 & 0.06 & 0.00 & 0.00 & 15.71\\
\cdashline{2-14}
& RTN & 3 & 16 & 128 & 35.24 & 31.35 & 51.32 & 8.95 & 3.80 & 2.91 & 1.00 & 1.22 & 15.90\\
& AWQ & 3 & 16 & 128 & 50.83 & 48.59 & 68.75 & 43.82 & 19.80 & 4.77 & 11.00 & 10.37 & 30.50\\
& GPTQ & 3 & 16 & 128 & 25.99 & 24.96 & 39.14 & 2.50 & 1.20 & 0.94 & 0.00 & 0.00 & 9.61\\
& GPTAQ & 3 & 16 & 128 & 24.87 & 23.33 & 36.84 & 1.90 & 2.20 & 1.07 & 0.00 & 0.00 & 10.17\\
\cdashline{2-14}
& RTN & 2 & 16 & 128 & 25.12 & 23.40 & 37.17 & 0.00 & 0.00 & 0.00 & 0.00 & 0.00 & 11.28\\
& AWQ & 2 & 16 & 128 & 25.09 & 25.41 & 36.51 & 0.00 & 0.00 & 0.10 & 0.00 & 0.00 & 10.54\\
& GPTQ & 2 & 16 & 128 & 25.13 & 23.18 & 36.51 & 0.00 & 0.00 & 0.02 & 0.00 & 0.00 & 8.50\\
& GPTAQ & 2 & 16 & 128 & 24.49 & 24.00 & 36.51 & 0.00 & 0.00 & 0.02 & 0.00 & 0.00 & 8.87\\
& SliM-LLM & 2 & 16 & 128 & 25.03 & 27.49 & 36.51 & 0.08 & 0.00 & 0.03 & 0.00 & 0.00 & 10.54\\
\cdashline{2-14}
& RTN & 1 & 16 & 128 & 25.26 & 23.03 & 36.51 & 0.00 & 0.00 & 0.00 & 0.00 & 0.00 & 0.00\\
& AWQ & 1 & 16 & 128 & Failed & Failed & Failed & Failed & Failed & Failed & Failed & Failed & Failed\\
& GPTQ & 1 & 16 & 128 & 25.26 & 23.03 & 36.51 & 0.00 & 0.00 & 0.00 & 0.00 & 0.00 & 0.00\\
& GPTAQ & 1 & 16 & 128 & 25.26 & 23.03 & 36.51 & 0.00 & 0.00 & 0.00 & 0.00 & 0.00 & 0.00\\
& BiLLM & 1 & 16 & 128 & 24.89 & 23.25 & 36.51 & 0.00 & 0.40 & 0.92 & 0.00 & 0.00 & 11.28\\
\hline
\end{tabular}
\end{adjustbox}
\label{tab:openpangu-7b-extended}
\end{table*}

\clearpage

\subsection{Visualization and Granularity Ablation}

\begin{figure*}[t]
    \centering
    \includegraphics[width=0.95\linewidth]{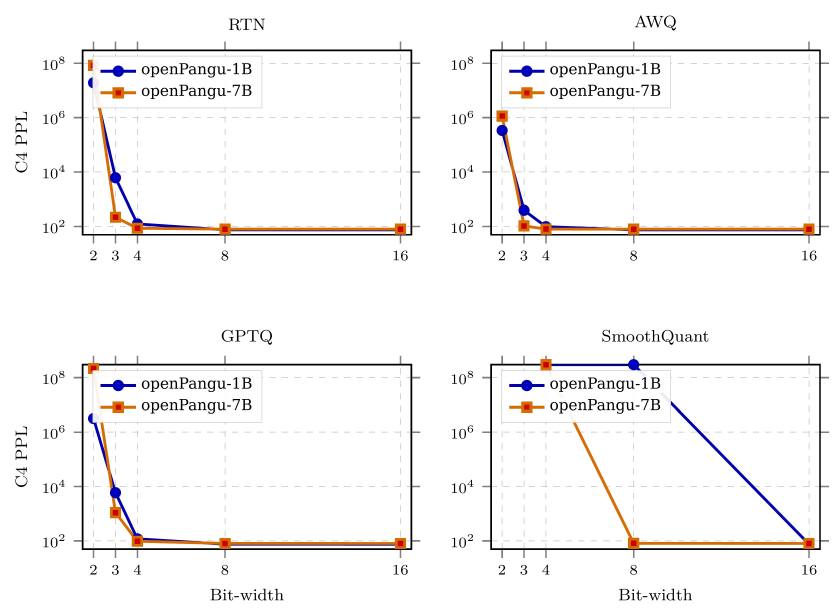}
    \caption{C4 perplexity of representative openPangu PTQ settings as a function of bit-width. RTN, AWQ, and GPTQ use per-group weight-only quantization with group size 128; for SmoothQuant, the 4-bit and 8-bit points denote W4A4 and W8A8 weight-activation quantization rather than weight-only quantization. Non-finite C4 values are shown at the axis ceiling.}
    \label{fig:openpangu-c4-ppl}
\end{figure*}

\begin{figure*}[t]
    \centering
    \includegraphics[width=0.95\linewidth]{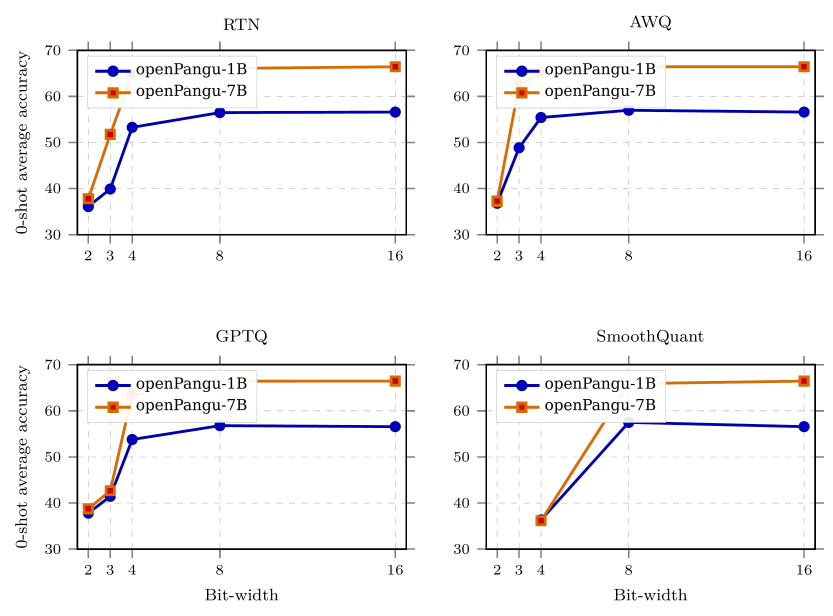}
    \caption{Zero-shot commonsense reasoning accuracy averaged over PIQA, ARC-Easy, ARC-Challenge, HellaSwag, Winogrande, and BoolQ as a function of bit-width for representative openPangu PTQ settings. RTN, AWQ, and GPTQ are weight-only settings, while SmoothQuant uses W4A4 or W8A8 weight-activation quantization.}
    \label{fig:openpangu-zeroshot-avg}
\end{figure*}

Figure~\ref{fig:openpangu-c4-ppl} and Figure~\ref{fig:openpangu-zeroshot-avg} visualize the main C4 perplexity and zero-shot commonsense trends from Table~\ref{tab:openpangu-1b-main} and Table~\ref{tab:openpangu-7b-main}. Each panel fixes a quantization method, the x-axis shows bit-width, and each curve corresponds to one model scale. For SmoothQuant, the 4-bit and 8-bit points denote W4A4 and W8A8, respectively. Non-finite C4 perplexity values are clipped to the top of the axis for visualization.

To compare per-channel and per-group scaling on openPangu, we additionally run a granularity ablation over all transformer linear weights in openPangu-1B and openPangu-7B. We first compare symmetric integer quantization with per-channel scaling against per-group scaling with group size 128 from 2 to 8 bits, using normalized weight reconstruction error as a hardware-independent proxy. We then verify the trend with NPU-based lm-eval on a representative task-level setting, openPangu-1B RTN-W4A16.

\begin{figure*}[t]
    \centering
    \includegraphics[width=0.90\linewidth]{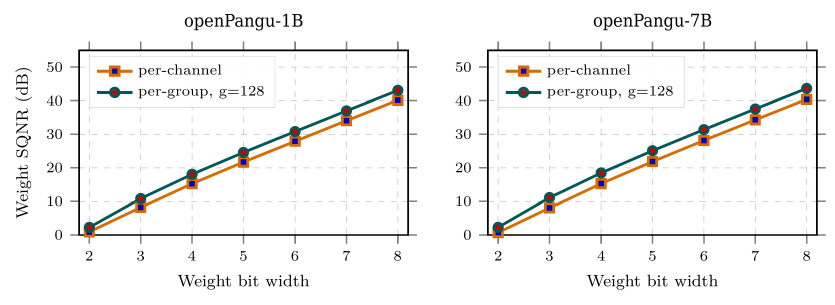}
    \caption{Weight reconstruction SQNR comparison between per-channel and per-group symmetric quantization on openPangu transformer linear layers. Higher is better.}
    \label{fig:openpangu-granularity-sqnr}
\end{figure*}

\begin{table}[t]
    \centering
    \small
    \caption{Per-bit granularity preference from the weight-error ablation. Values in parentheses are the SQNR gain of per-group over per-channel.}
    \begin{adjustbox}{max width=\linewidth}
    \begin{tabular}{lcccc}
    \hline
    \textbf{Model} & \textbf{W8} & \textbf{W4} & \textbf{W3} & \textbf{W2} \\
    \hline
    openPangu-1B & per-group (2.92 dB) & per-group (2.80 dB) & per-group (2.61 dB) & per-group (1.34 dB) \\
    openPangu-7B & per-group (3.24 dB) & per-group (3.19 dB) & per-group (3.05 dB) & per-group (1.57 dB) \\
    \hline
    \end{tabular}
    \end{adjustbox}
    \label{tab:openpangu-granularity-preference}
\end{table}

\begin{table}[t]
    \centering
    \small
    \caption{NPU lm-eval comparison of per-channel and per-group RTN-W4A16 on openPangu-1B.}
    \begin{adjustbox}{max width=\linewidth}
    \begin{tabular}{lccc}
    \hline
    \textbf{Granularity} & \textbf{WikiText2 PPL} $\downarrow$ & \textbf{PIQA Acc.} $\uparrow$ & \textbf{PIQA Acc. Norm.} $\uparrow$ \\
    \hline
    per-channel & 140.09 & 61.10 & 60.99 \\
    per-group, $g=128$ & 43.30 & 64.64 & 64.58 \\
    \hline
    \end{tabular}
    \end{adjustbox}
    \label{tab:openpangu-granularity-lm-eval}
\end{table}

The weight-error ablation consistently favors per-group scaling for both model sizes across W2--W8. The gain is modest but non-negligible at W8 and grows to around 2.6--3.2 dB in the practically relevant W3--W7 range. Table~\ref{tab:openpangu-granularity-lm-eval} shows that this trend transfers to task-level evaluation: at RTN-W4A16, per-channel scaling raises WikiText2 PPL from 43.30 to 140.09 and drops PIQA accuracy from 64.64 to 61.10. Combined with the accuracy tables, these results suggest that weight-only and low-bit settings should use per-group scaling whenever the deployment kernel supports it; W8 per-channel scaling can be treated as a compatibility fallback; and per-group weight scaling alone is insufficient to compensate for W4A4 activation quantization, where activation error dominates.

\subsection{Task-Level Observations}

\textbf{Model size matters.}
The 7B model has enough redundancy to make weight-only W4 quantization practical, while the 1B model does not show the same margin. On openPangu-7B, AWQ-W4 keeps C4 PPL almost unchanged relative to FP16 (80.60 vs. 80.43), and its zero-shot commonsense average only drops from 66.43 to 65.15. RTN-W4 is similar on the commonsense average (64.87) and even keeps BoolQ slightly above the FP16 score. In contrast, openPangu-1B suffers a sharper capability shift at W4: AWQ-W4 still has a finite C4 PPL of 98.98 and a commonsense average of 55.41, but GSM8K falls from 30.10 to 10.01, MATH-500 from 23.20 to 11.20, MBPP from 21.60 to 6.20, and HumanEval from 20.12 to 9.15. This indicates that commonsense classification can remain deceptively stable after W4 compression, while mathematical and coding abilities expose the reduced capacity of the smaller model.

\textbf{Perplexity is necessary but not sufficient.}
Some W4 settings preserve perplexity better than they preserve downstream capabilities. For example, openPangu-1B AWQ-W4 has the best W4 perplexity among the 1B weight-only methods, but its CMMLU and C-Eval scores are lower than GPTQ-W4, and its math/code scores remain weak. Conversely, openPangu-7B RTN-W4 does not lead on C4 PPL, but it is competitive on MMLU, GSM8K, and MATH-500. The same pattern appears at higher bit-widths: several W5--W6 results slightly exceed FP16 on individual tasks, such as openPangu-7B RTN-W5 on MMLU and RTN-W6 on GSM8K, while other metrics remain close to or below FP16. These improvements should be interpreted as task-level variation rather than as evidence that quantization universally improves the model. The practical lesson is that perplexity can screen out clearly broken settings, but deployment selection still needs the downstream task family that matches the intended workload.

\textbf{Method ranking changes with bit-width.}
At W8, most weight-only methods are effectively indistinguishable from FP16 on the main metrics, so implementation simplicity may matter more than small score differences. At W4, method choice becomes important: AWQ is the strongest weight-only choice for openPangu-7B in this suite, with C4 PPL close to FP16 and the best W4 commonsense average, while GPTQ and GPTAQ have larger PPL increases and weaker math/code behavior. At W3, only AWQ keeps openPangu-7B in a partially usable regime: AWQ-W3 reaches a 62.46 commonsense average, while RTN-W3 drops to 51.74 and GPTQ/GPTAQ-W3 collapse to around 43. For openPangu-1B, even AWQ-W3 is not a general-purpose setting, despite being much better than the other W3 methods. This suggests that activation-aware clipping is especially useful at the edge of feasible precision, but it cannot fully compensate for insufficient model capacity.

\textbf{Ultra-low precision has a clear failure signature.}
The W2, binary, and SliM-LLM settings show a consistent pattern: perplexity becomes extremely large or non-finite, commonsense scores approach chance-level behavior, and math/code scores are nearly zero. The pattern is visible for both model sizes, although openPangu-7B BiLLM retains a higher BoolQ score than most other binary settings. These isolated survivals do not change the overall conclusion, because the same rows fail on perplexity and broad downstream metrics. Ultra-low-bit deployment therefore needs additional recovery mechanisms before it can be treated as a default option for openPangu.

\textbf{Granularity affects real task behavior, not only reconstruction error.}
The weight-error ablation shows that per-group scaling improves SQNR over per-channel scaling across both model sizes and all tested bit-widths. More importantly, the NPU lm-eval check confirms that this is not only a proxy metric: for openPangu-1B RTN-W4A16, per-channel scaling increases WikiText2 PPL from 43.30 to 140.09 and reduces PIQA accuracy from 64.64 to 61.10. This makes granularity a deployment-level design choice rather than a cosmetic quantizer detail. When a low-bit Ascend kernel supports group-wise scaling, the accuracy evidence favors using it.

\textbf{W+A quantization requires caution.}
The SmoothQuant-W8A8 result is largely usable on 7B and moderately stable on 1B, but W4A4 is not. openPangu-7B SmoothQuant-W8A8 keeps C4 PPL close to FP16 and maintains a 65.88 commonsense average, whereas W4A4 produces non-finite PPL and drops the commonsense average to 36.14. The 1B W8A8 run shows a non-finite C4 PPL while retaining moderate classification scores, which suggests that simulated activation quantization can produce metric-specific numerical failures before all downstream tasks collapse. Because these experiments simulate activation quantization during evaluation, the results should be treated as accuracy diagnostics rather than final kernel-level throughput claims.

\subsection{Discussion}

\textbf{W8 weight-only quantization is the safest default.}
The main practical recommendation from these experiments is that W8 weight-only quantization is the low-risk setting for both openPangu models on the current Ascend NPU software stack. It preserves perplexity and broad benchmark accuracy while requiring only a conservative weight-only deployment path. For scenarios where accuracy regressions are difficult to debug after deployment, W8 should be preferred over more aggressive settings.

\textbf{W4 is a model- and workload-dependent choice.}
For tighter memory budgets, W4 is viable mainly for openPangu-7B. AWQ-W4 provides the best overall balance in the collected 7B runs, while RTN-W4 is a competitive fallback that needs no calibration data. GPTQ-W4 and GPTAQ-W4 are usable but weaker on several individual datasets, especially in the math and code portion of the suite. For openPangu-1B, W4 should be treated as task-specific: it may be adequate for lightweight commonsense classification, but it is not a safe general replacement for FP16 or W8 when math and code are important.

\textbf{Activation quantization is the main unresolved bottleneck.}
The contrast between weight-only W4 and SmoothQuant-W4A4 indicates that activation error is more damaging than weight compression alone in the tested flow. This is especially relevant for Ascend deployment, where throughput gains often require kernel-level support for both low-bit weights and efficient activation handling. The current results support weight-only compression as a near-term path, but they do not support W4A4 as a reliable accuracy-preserving setting.

\textbf{Ultra-low precision needs method-specific recovery.}
Ultra-low precision should not be used as a default deployment setting. The current W2, binary, and SliM-LLM 2-bit results are dominated by unstable perplexity and low task accuracy. AWQ-W3 on openPangu-7B is the one exception worth further engineering, but it still needs workload-specific validation and a real deployment-kernel check before it can be considered production-ready.

\textbf{Hardware-aware evaluation should separate accuracy and speed.}
The experiments are NPU-oriented but not full deployment benchmarks. They give an accuracy map under the current software stack, while actual production decisions also require packed kernels, memory measurements, throughput, and energy profiling. This separation is important: a quantization setting that is accurate in simulation may still be unattractive without an efficient kernel, and a fast kernel is not useful if it pushes the model into the collapse region observed at W2, binary, or W4A4.

\section{Conclusion}

We presented an empirical study of openPangu 1B and 7B quantization on Ascend 910B1 NPUs under a unified PTQ evaluation workflow. The main conclusions are consistent across tasks: W8 weight-only quantization is essentially lossless; W4 is practical for openPangu-7B, especially with AWQ, but significantly more damaging for openPangu-1B; and W2, binary, and W4A4 settings are not reliable under the current flows. These findings give a concrete, NPU-specific basis for choosing openPangu quantization settings and for prioritizing future work on activation quantization and ultra-low-bit recovery.

\section{Limitations and Future Work}

\textbf{Limitations.}
This study is intentionally accuracy-focused. Several experiments use simulated quantization rather than fully packed low-bit deployment kernels, and the SmoothQuant W+A runs simulate activation quantization during evaluation. We therefore do not claim NPU latency, throughput, or energy speedups from these tables. The results may also vary with model implementation details, calibration sample selection, random evaluation effects, and the PyTorch NPU/CANN compatibility state.

\textbf{Future work.}
Future work should complement this accuracy study with real Ascend deployment kernels, including packed weight formats, memory measurements, and end-to-end generation throughput. This is especially important for distinguishing settings that are merely accurate in simulated or fake-quantized evaluation from settings that provide measurable system-level benefits on the target NPU stack. Another direction is to revisit the promising openPangu-7B AWQ-W3 result with stronger calibration selection, channel reordering~\cite{yuan2023rptq}, and rotation-based quantization~\cite{liu2024spinquant}. The current results suggest that W3 is close to the boundary between usable and collapsed behavior, so additional recovery techniques should be evaluated with both perplexity and downstream task metrics. Finally, the gap between weight-only W4 and W4A4 indicates that activation outliers and numerical stability are major obstacles for aggressive deployment; future work should therefore combine accuracy evaluation with activation-aware kernels and runtime checks for non-finite behavior.

\end{document}